\documentclass[letterpaper, 10pt, conference]{ieeeconf}  

\usepackage{times}
\usepackage{epsfig}
\usepackage{graphicx}
\usepackage{amsmath}
\usepackage{amssymb}
\usepackage[pagebackref=true,breaklinks=true,letterpaper=true,colorlinks,bookmarks=false, breaklinks=true]{hyperref}
\usepackage{breakurl}
\usepackage{mathptmx}
\usepackage{amsopn}
\usepackage{algorithm}
\usepackage{algorithmic}
\usepackage{multirow}
\usepackage{epstopdf}
\usepackage[caption=false]{subfig}
\usepackage[flushleft]{threeparttable}

\usepackage{hhline}

\usepackage[T1]{tipa}
\usepackage{siunitx}
\usepackage[T1]{fontenc}

\IEEEoverridecommandlockouts                              
\title{\LARGE \bf
ExpressionBot: An Emotive Lifelike Robotic Face for Face-to-Face Communication
}

\author{Ali Mollahosseini, Gabriel Graitzer, Eric Borts, Stephen Conyers, \\
Richard M. Voyles, Ronald Cole, and Mohammad H. Mahoor
\thanks{*This work is partially supported by the NSF grants IIS-1111568, IIS- 1111544, and CNS-1427872.}
\thanks{A. Mollahosseini, Gabriel Graitzer, Stephen Conyers and Mohammad H. Mahoor are with the department of Electrical and Computer Engineering at the University of Denver, CO 80210. e-mails: \tt\small{{ali.mollahosseini, gabriel.graitzer, stephen.conyers,  mmahoor}@du.edu.}}
\thanks{Richard M. Voyles is with the College of Technology at Purdue University, West Lafayette, IN 47907. E-mail: {\tt\small rvoyles@purdue.edu}}
\thanks{Eric Borts and Ronald Cole are with the Boulder Language Technologies, Boulder CO 80301. e-mails: \tt\small{{eborts, rcole}@bltek.com}}
}

\begin{document}

\maketitle
\thispagestyle{empty}
\pagestyle{empty}

\begin{abstract}
This article proposes an emotive lifelike robotic face, called ExpressionBot, that is designed to support verbal and non-verbal communication between the robot and humans, with the goal of closely modeling the dynamics of natural face-to-face communication.  The proposed robotic head consists of two major components: 1) a hardware component that contains a small projector, a fish-eye lens, a custom-designed mask and a neck system with 3 degrees of freedom; 2) a facial animation system, projected onto the robotic mask,  that is capable of presenting facial expressions, realistic eye movement, and accurate visual speech. We present three studies that compare Human-Robot Interaction with Human-Computer Interaction with a screen-based model of the avatar.  The studies indicate that the robotic face is well accepted by users, with some advantages in recognition of facial expression and mutual eye gaze contact.
\end{abstract}

\section{Introduction}
\label{intro}
Although robots are finding their place in our society as artificial pets, entertainers, and tools for therapists, current technologies have yet to reach the full emotional and social capabilities necessary for rich and robust interaction with human beings. To achieve this potential, research must imbue robots with the emotional and social capabilities -both verbal and non-verbal- necessary for rich and robust interaction with human beings. This article describes research in which robotic heads can model natural face-to-face communication with individuals. Human face-to-face communication is based on multiple communication channels including auditory and visual. Facial expressions and gaze direction are the most important visual channels in human face-to-face communication. These channels of communications are considered in our research. 

Despite significant progress towards development of realistic robotic heads, a number of problems remain to be solved. Social robots such as Paro~\cite{_paro} have the robustness and cost effectiveness for large scale, but lack the sophistication for deep social interaction. Other robots such as Kismet~\cite{_kismet} exhibit facial expressions and head and ear movement using mechanical components, however, once these mechanical platforms are built, they are fixed and cannot be readily modified. On the other hand, more human-like robots such as Simon~\cite{_simon} possess profound capabilities for social interaction, but due to large number of actuators in their mechatronic faces, they are expensive and maintenance-intensive for large-scale trials. Another potential problem in the design of robotic heads is the ``Uncanny Valley'' effect~\cite{mori_uncanny_2012}, where the effect of aesthetic design of a robot may influence the user's experience, perception, and acceptance of the robot. 

Given the tremendous effort required to develop robot heads and the number of design choices that must be made, including aesthetic face design, mechanical design and construction, hardware implementation, etc, it is often difficult to redesign or rebuild the head based on user experiences and limitations discovered during research. An alternative approach that overcomes many of these problems is to use state-of-the-art character animation technologies to create 3D avatar models that produce natural speech and facial expressions, and project these models onto a robotic face that can move like a human head. Projecting an animated facial character on a 3D human-like head has several advantages over displaying it on a 2D flat screen. The first advantage is the physical embodiment and movement of the 3D head that makes the robot be more believable and better perceived by the users~\cite{wainer_role_2006}. The other advantage is perception of mutual eye gaze which plays a vital role in face-to-face communication. It is known that, the perception of 3D objects that are displayed on 2D surfaces are influenced by the Mona Lisa effect~\cite{todorovic_geometrical_2006}. In other words, the orientation of the object in relation to the observer will be perceived as constant regardless of observer's position~\cite{Delaunay_2010_study}. For instance, if the 2D projected face is gazing forward, mutual gaze is perceived with the animation, regardless of where the observer is standing/sitting in relation to the display. This effect is significantly reduced by projecting the face on a 3D human-like head. 

This article describes our current progress towards designing, manufacturing, and evaluating a lifelike robotic head, called ExpressionBot, with the capability of showing facial expressions, visual speech, and eye gaze. Our eventual goal is to provide the research community with a low-cost portable facial display hardware equipped with a software toolkit that can be used to conduct research leading to a new generation of robotic heads that model the dynamics of face-to-face communication with individuals in different social, learning, and therapeutic contexts. 

The remainder of this paper is organized as follows. Section~\ref{sec:RelatedWork} reviews the related work on the design of social robotic faces. Section~\ref{sec:Mechanism and Design} presents the design, development, and production of the proposed robotic face. Section~\ref{sec:HRI Experiments and Results} presents the results of our pilot studies on user experiences with the robotic face, while Section~\ref{sec:Conclusion} presents our conclusions.  

\section{Related Work}
\label{sec:RelatedWork}
There are many designs for robotic faces ranging from 2D graphical avatars to mechanically controlled robotic faces. These designs fit into four main categories: Mechatronic Faces, Android Faces, Onscreen Avatars, and light-projected (retro-projected) physical Avatars. These categories are discussed in the following.

Mechatronic robotic faces are physically implemented robots that use mechatronic devices and electric actuators to control facial elements. Kismet~\cite{_kismet} is one the first and famous expressive mechatronic robots with many features such as eye lids, eye brows, lips and even expressive ears. Another example is the Philips iCat~\cite{van_breemen_icat_2005} which has a cat-like head and torso with mechanical lips, eye lids, and eye brows. Mechatronic robotic faces have the advantage of being 3D, but they are inflexible, unrealistic, and have limited ability to display facial expressions and speech. These faces look very much like a stereotypical robot rather than a human face. 

Android faces are other physically implemented robots that are originated from Animatronics. They have a larger number of mechatronic actuators controlling a flexible elastic skin; therefore they look more realistic and seem more like a human rather than a robot. Example of android faces are Albert Hubo~\cite{hanson_expanding_2005}, HRP-4C~\cite{_successful}, and Zeno~\cite{hanson_zeno_2009}. It is an interesting research question as to whether android faces that closely model human looks and behaviors will enter the uncanny valley as their realism mimics humans. Due to larger number of actuators and their interaction with skin, they look more expressive than mechatronic faces. However, they are mechanically very complex, expensive to design, build and maintain. 

The on screen avatar class, such as Grace~\cite{gockley_grace_2004} and Second Life~\cite{_second_life} are the simplest and earliest robotic faces. Animations for these models can be made by developing a model for each expression, morphing between them, and then rendering the result to a computer screen. Despite their low cost and high flexibility, they naturally have several limitations due to using a flat display as an alternative to a three dimensional physical head. For example, aside aesthetic unpleasantness, they suffer from lack of establishing mutual gaze (due to Mona Lisa effect) and physical embodiment that both play vital roles in natural and realistic face-to-face communication. 

The final category, and the focus of our research, is the light-projected physical avatars; these consist of translucent 3D masks with 2D/3D avatars projected onto them. Since the robotic face is projected onto a mask, the robotic face can range from cartoon-like to photorealistic. Light-projected physical avatars are thus a highly flexible research tool, relative to mechatronic and android faces. Factors such as engagement, embodiment, believability, credibility and realism can be investigated based on the appearance and behaviors of the 3D animated models, and the robotic head and neck movement. Moreover, such a system can avoid the Mona Lisa effect~\cite{todorovic_geometrical_2006} and hence users can correctly perceive the robot's eye gaze direction.  Additional features of robotic avatars include relatively low development cost, low power consumption, potentially low weight and fast reaction. 

One of the early examples of light-projected physical avatars is the Dome robot~\cite{hashimoto_facial_2005} where a cartoonish animated face is projected onto a dome-shaped mask. The dome mask makes the image and display calibration process easy. However, it lacks human face realism and the results appear cartoonish. The Lighthead robotic face~\cite{Delaunay_2009} is another example that also projects an animation onto a face-shaped translucent mask, resulting in a more realistic appearance. It is capable of displaying a wide range of facial expressions and emotions. In~\cite{kuratate_mask_bot_2011} the authors presented a mask head robot, called Mask-bot, that generates visual speech, head movements, facial expressions and eye movements. Then later in~\cite{pierce_development_2012}, Mask-bot 2i was introduced with an automatic approach for projection calibration by using a series of gray-coded patterns in a calibration booth which supports interchangeable masks. Both Mask-bot and Mask-bot 2i use human talking head animation that is photo realistic, which is not as flexible as computer animation.

Furhat robot~\cite{al_moubayed_furhat_2012} is another human-like light-projected robotic head that utilizes computer animation to deliver facial movements. In~\cite{al_moubayed_furhat_2012}, the designers of Furhat had also a study on perception of animation's gaze on 3D projected against flat screens and demonstrated the limitations of flat screens in delivering accurate direction of gaze due to Mona Lisa effect, which limits having situated, multiparty interaction in onscreen Avatars. Furhat uses a pan-tilt unit for the neck system which has only 2 DOF pitch and yaw. Also Furhat uses a mirror instead of a fish eye lens which makes the robot to have a larger form factor. 

Socibot~\cite{SociBot} is a commercialized retro-projected social robotic head. It has an advanced neck system with 3 DoFs with force/torque sensory measurement that can capture and replicate the head movement trajectory. It uses a universal mask to show different avatar characters. Compared to our system, the quality of animation and facial expression generation are low. The visual speech is also na\"{\i}ve (i.e. open and close mouth). Moreover, their system is equipped with a speech synthesizer but it is not able to align recorded voices with accurate visual speech.

\section{Mechanism and Design}
\label{sec:Mechanism and Design}
The ExpressionBot consists of three main sections, the neck control system, the display system and the animation application. The neck system controls the projector and mask position allowing it to be rotated by the application to track faces and head gestures. The display system consists of a small projector with a fish eye lens that projects the animation on a human like (head shaped) face mask. The animation application displays a face animation along with speech and emotion to be projected on to the mask.

\subsection{Neck System}
\label{sec:Neck System}
The neck mechanism of our existing prototype has three degrees of freedom (DoF) providing a total of 150$^{\circ}$ of yaw (x-y plane), 30$^{\circ}$ of pitch (x-z plane) and 30$^{\circ}$ of roll (y-z plane). Solidworks CAD tool was used to design and maximize the range of motion, resulting in a light, compact, and quiet mechanism. These design constraints were achieved using a 6''$\times$6'' footprint, low friction plastic gears and brushless servomotors. Also, the small footprint allows the neck and projector to be easily shrouded by the mask, and allows the user to control the distance from the mask to the lens in order to project the clearest possible image.

\subsection{Display System}
\label{sec:Display System}
Our system uses a Dell DLP M110 portable projector. The projector is capable of up to 300 ANSI Lumens under normal indoor illumination conditions, can display a maximum resolution of 1200$\times$800, and has a 10000:1 contrast ratio. Attached to the projector is a Nikon Fisheye Converter FC-E8 which provides a viewing angle of approximately 183 degrees. This allows the projector to display to the whole mask from a relatively close distance.

\begin{figure}[h]
\centering
\subfloat[]
{
	\centering
    \includegraphics[scale = 0.16]{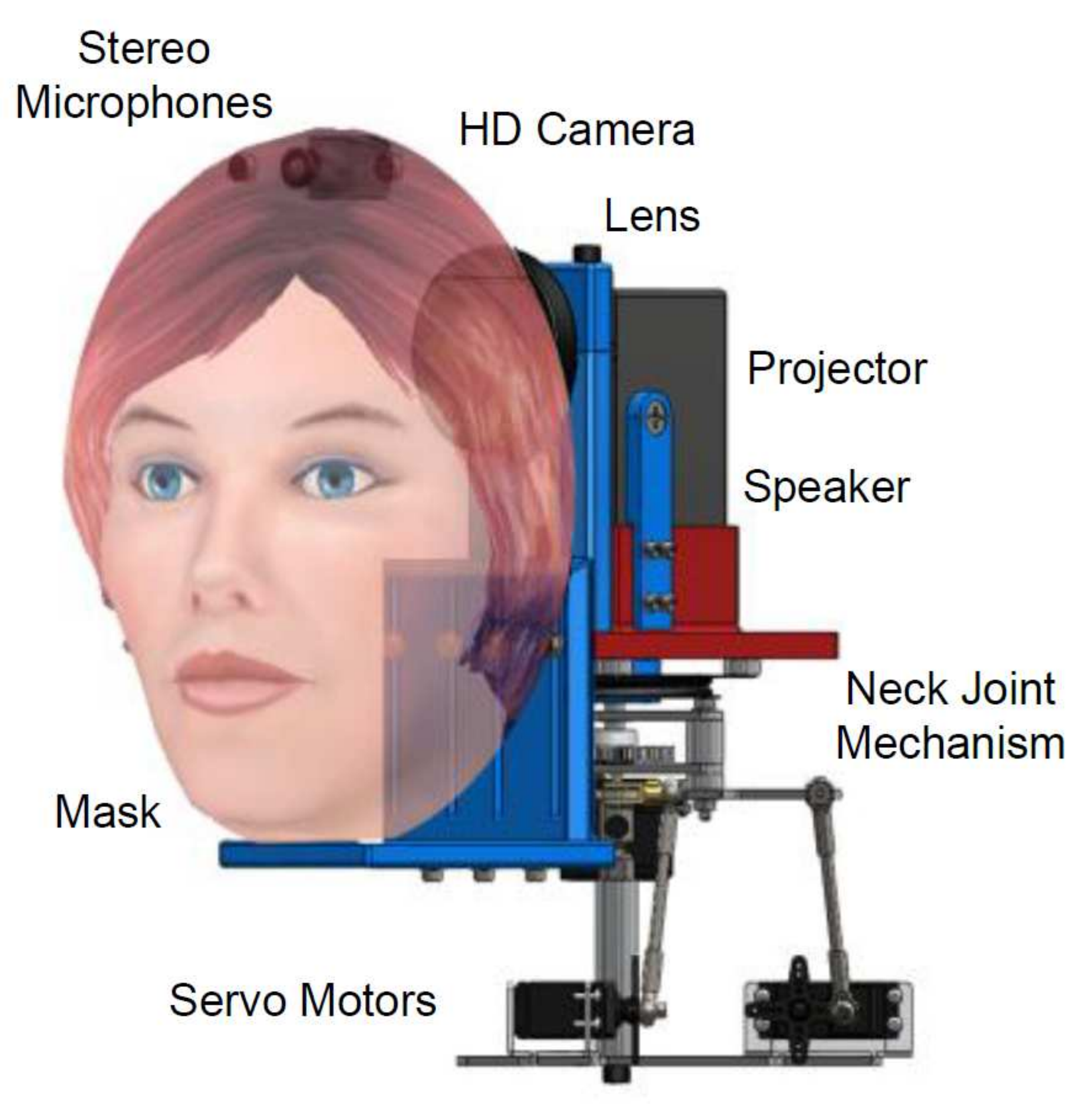}
}
\subfloat[]
{
    \includegraphics[scale = 0.18]{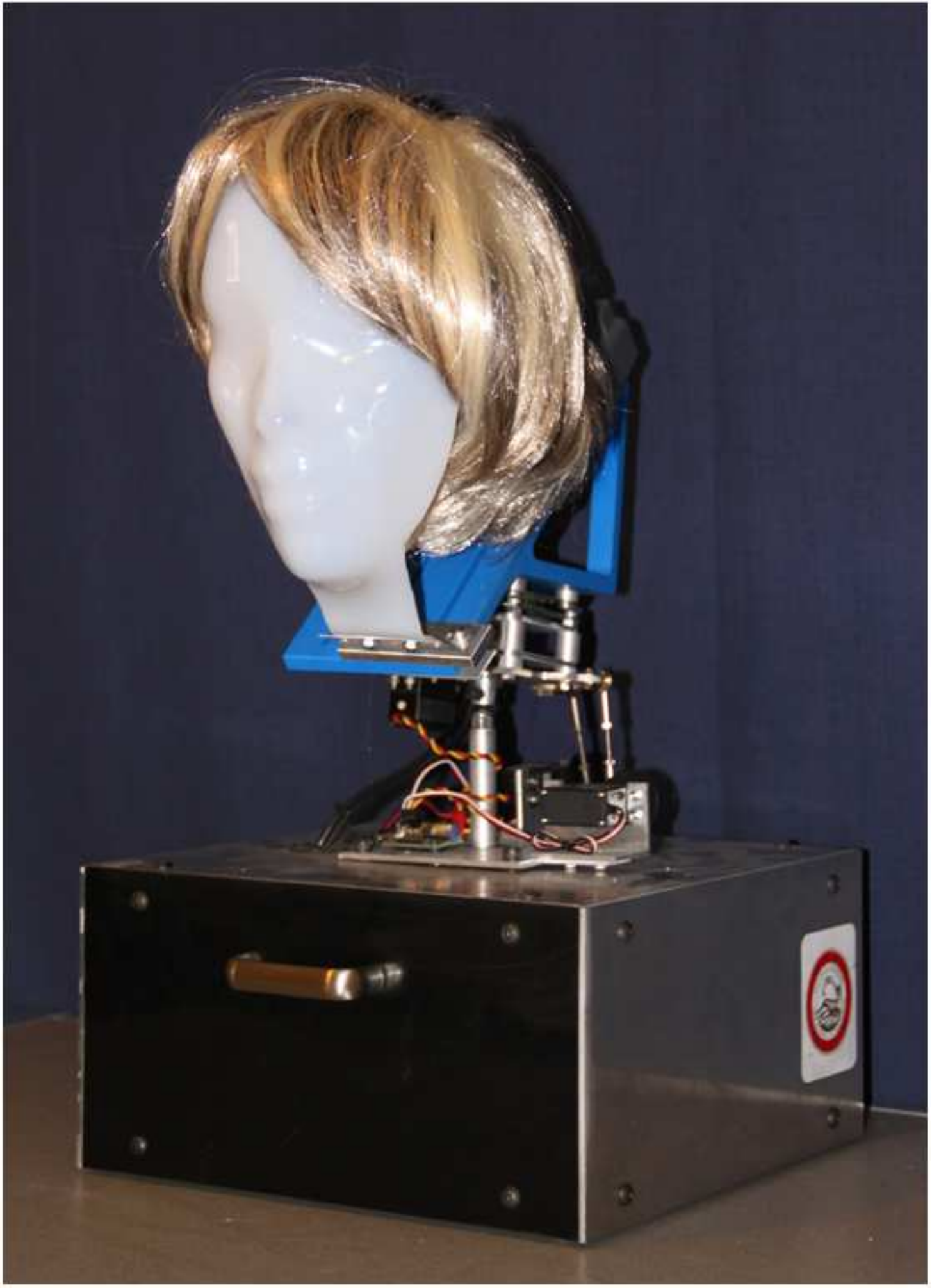}
}
\caption{\label{fig:Proposed robot system}
ExpressionBot's design and configuration.
}
\vspace{-0.2cm}
\end{figure}

To create the mask we designed a mold using the 3D model of the neutral face in Autodesk Maya. We 3D printed this mold and used it to vacuum form a 1/8 inch sheet of white translucent acrylic plastic. Then, we added a metal band from top of the mask to the projector, which allows us to mount a wig on the robot's head. This makes the ExpressionBot more aesthetically pleasant and natural, and covers the lights coming out from the sides of the mask due to fish eye lens wide projection-angle (see Fig.~\ref{fig:Proposed robot system}).

\subsection{Animation}
\label{sec:Animation}
We developed a face animation in C\# .Net for accurate natural visual speech and show expression based on multi-target morphing method~\cite{ma_animating_2004}. Recorded utterances are processed by the Bavieca speech recognizer~\cite{bolanos_bavieca_2012}, which receives the sequence of words and the speech waveform as input, and provides a time-aligned phonetic transcription of the spoken utterance. The aligned phonemes are represented using the International Phonetic Alphabet (IPA), a standard that is used to provide a unique symbolic notational for the realization of phonemes in all of the world's languages~\cite{association_handbook_1999}. As IPA is intended as a standard for the phonemic and phonetic representation of all spoken languages, having IPA in our system will allow us to add other languages easily as long as the speech recognizer is trained for that language.

\begin{figure}
\centering
\subfloat[][\label{Neutral_face}Neutral face]
{
	\centering
    \includegraphics[scale = 0.15]{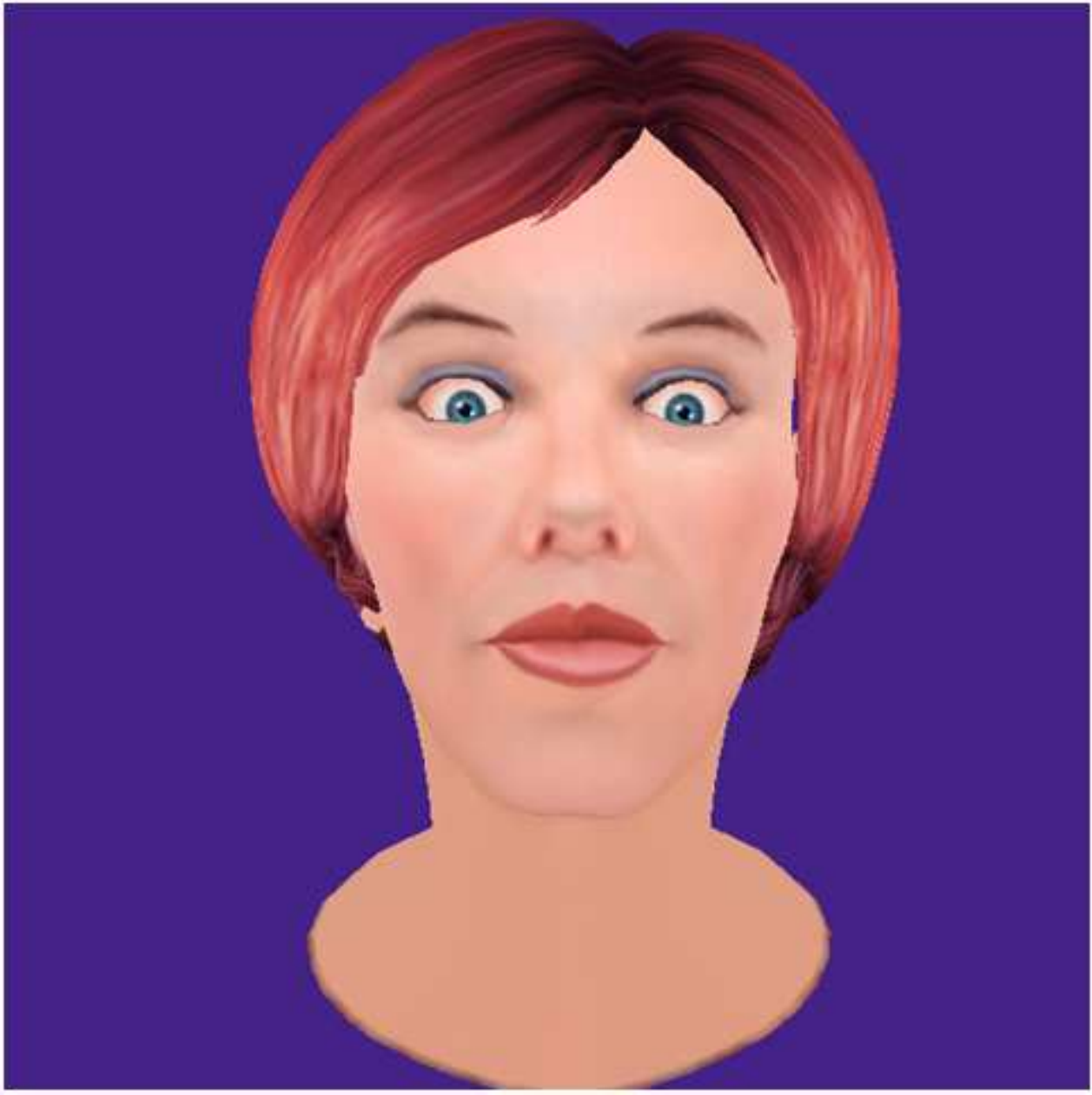}
}
\subfloat[][\label{Viseme_S} Viseme S]
{
	\centering
    \includegraphics[scale = 0.15]{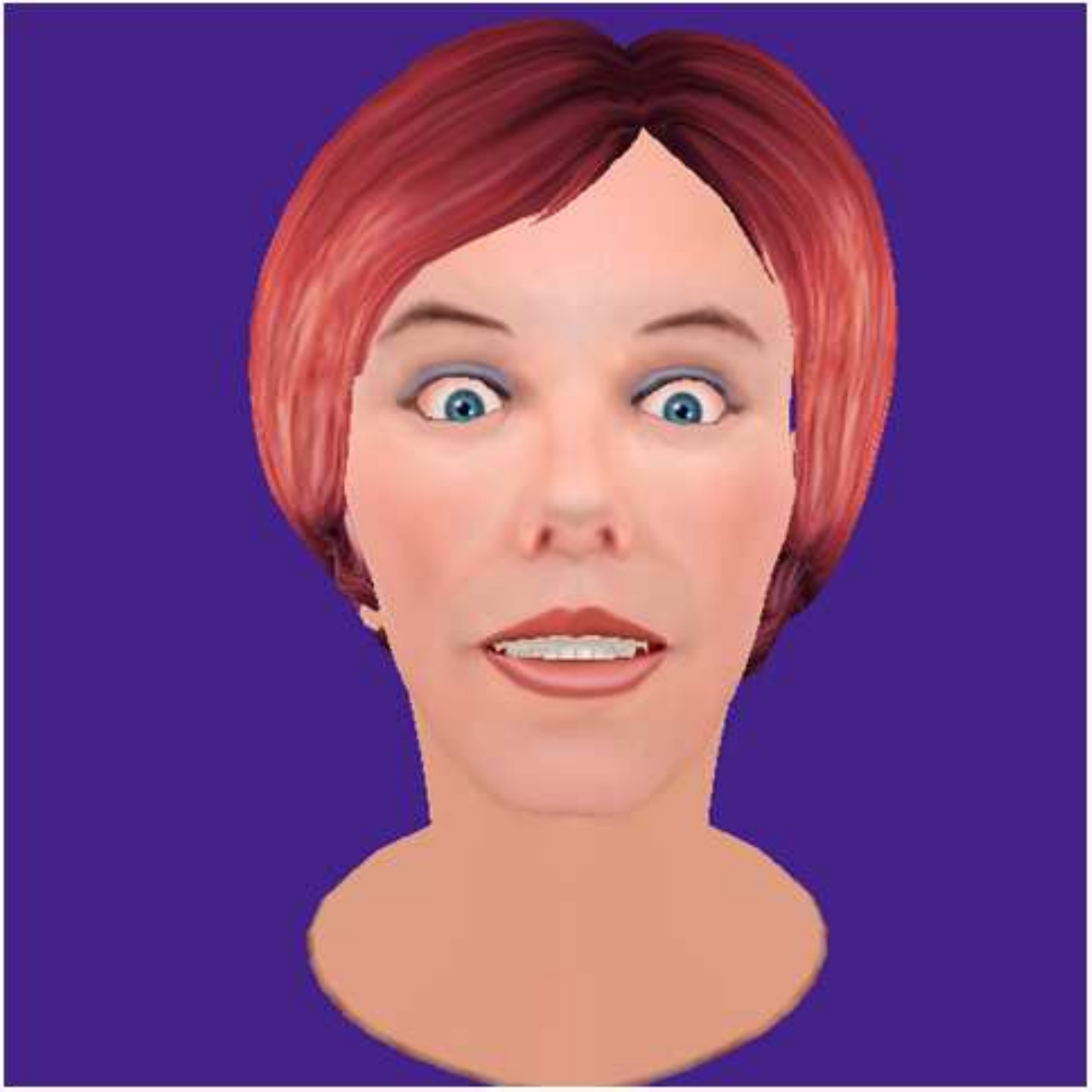}
}
\subfloat[][\label{Anger} Anger]
{
	\centering
    \includegraphics[scale = 0.15]{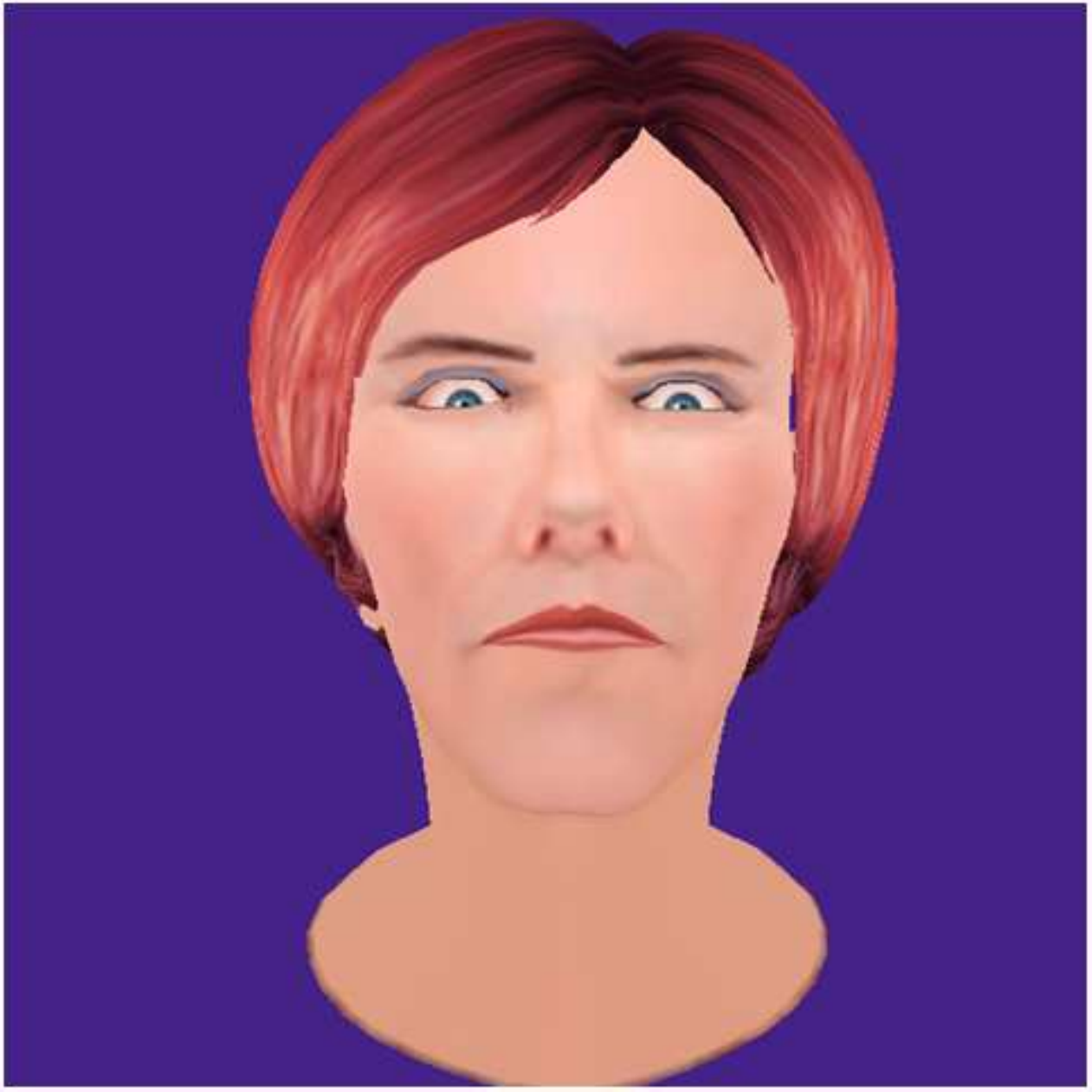}
}
\subfloat[][\label{disgust} Disgust]
{
	\centering
    \includegraphics[scale = 0.15]{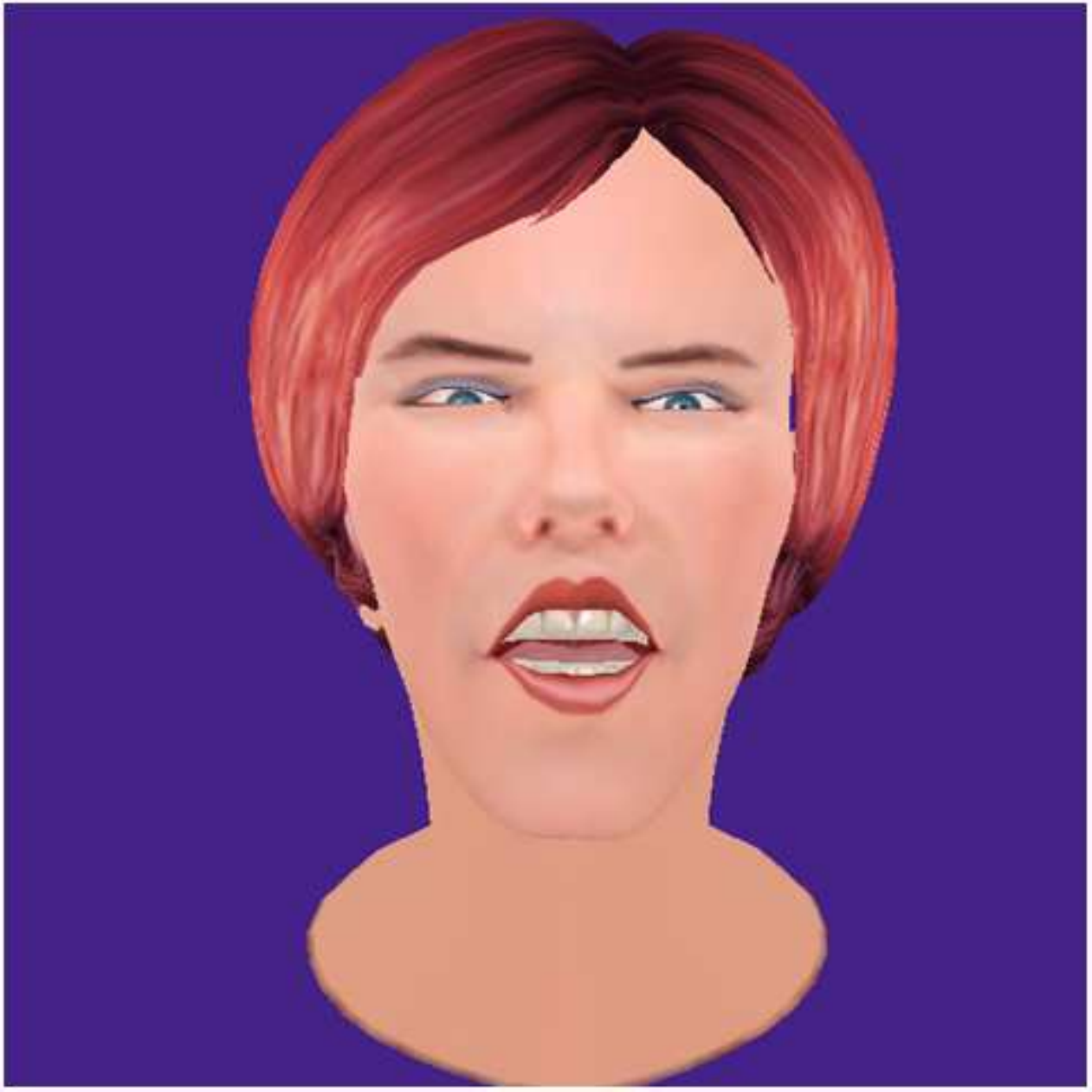}
}
\vspace{-0.3cm}
\subfloat[][\label{Fear} Fear]
{
	\centering
    \includegraphics[scale = 0.15]{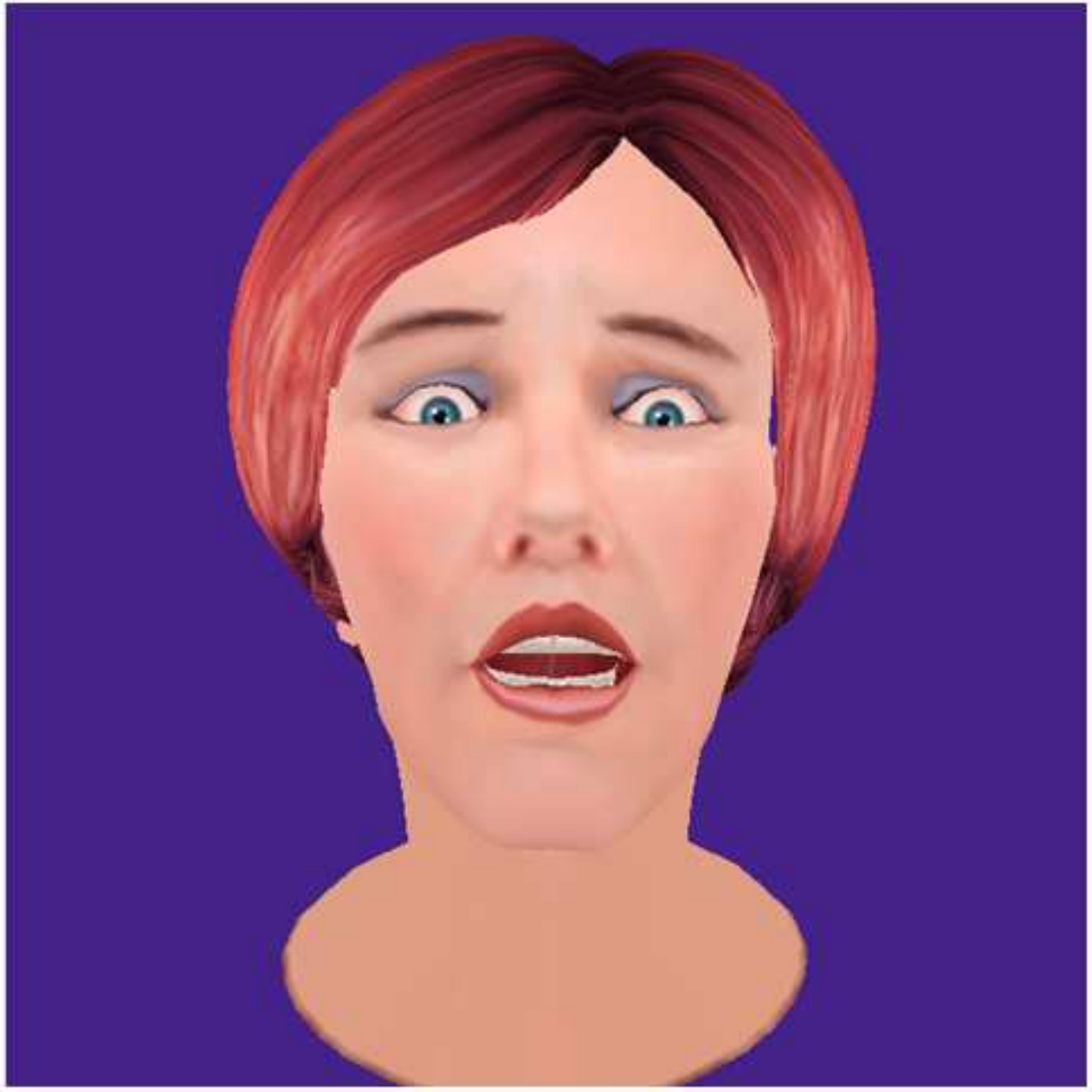}
}
\subfloat[][\label{joy}Joy]
{
	\centering
    \includegraphics[scale = 0.15]{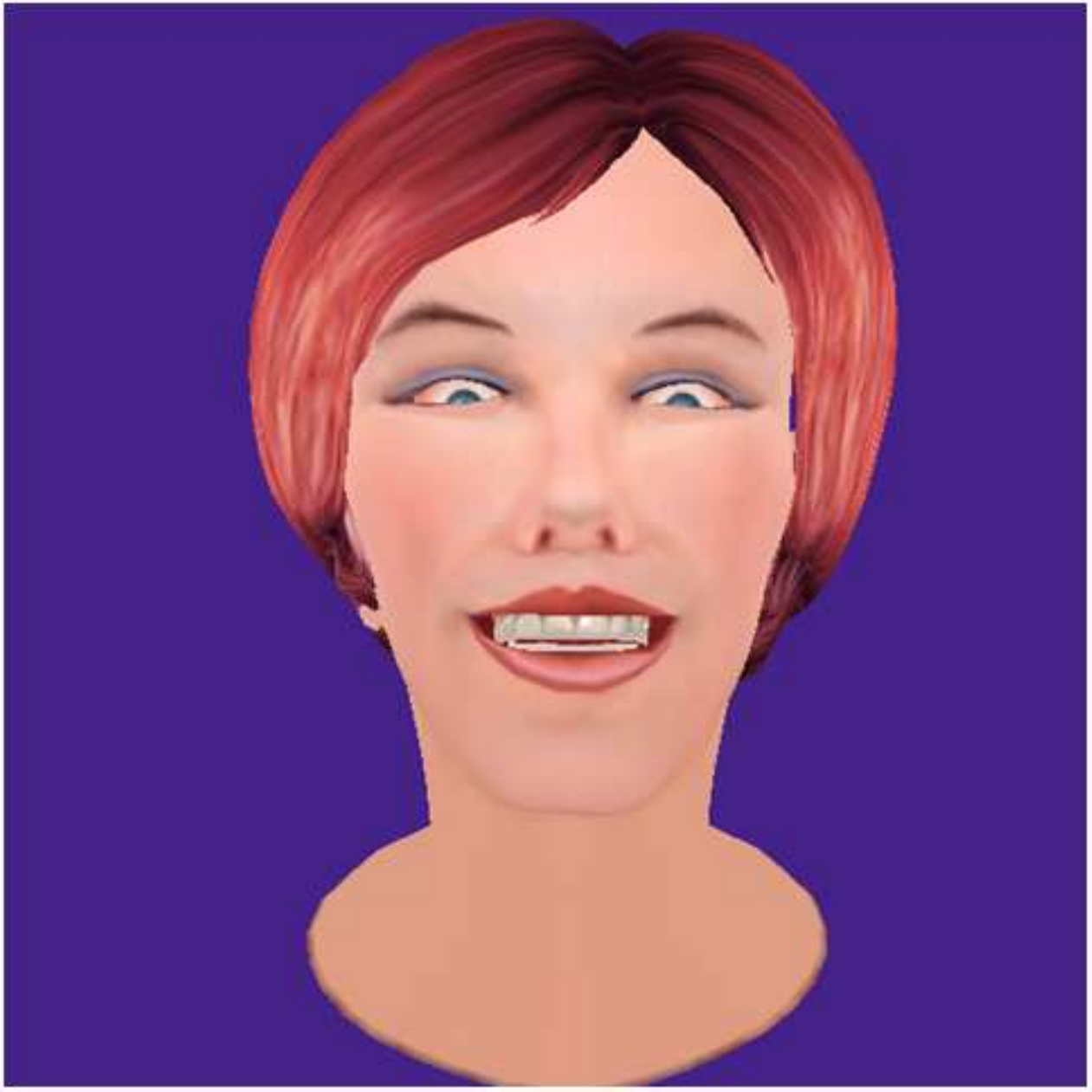}
}
\subfloat[][\label{sad}Sadness]
{
	\centering
    \includegraphics[scale = 0.15]{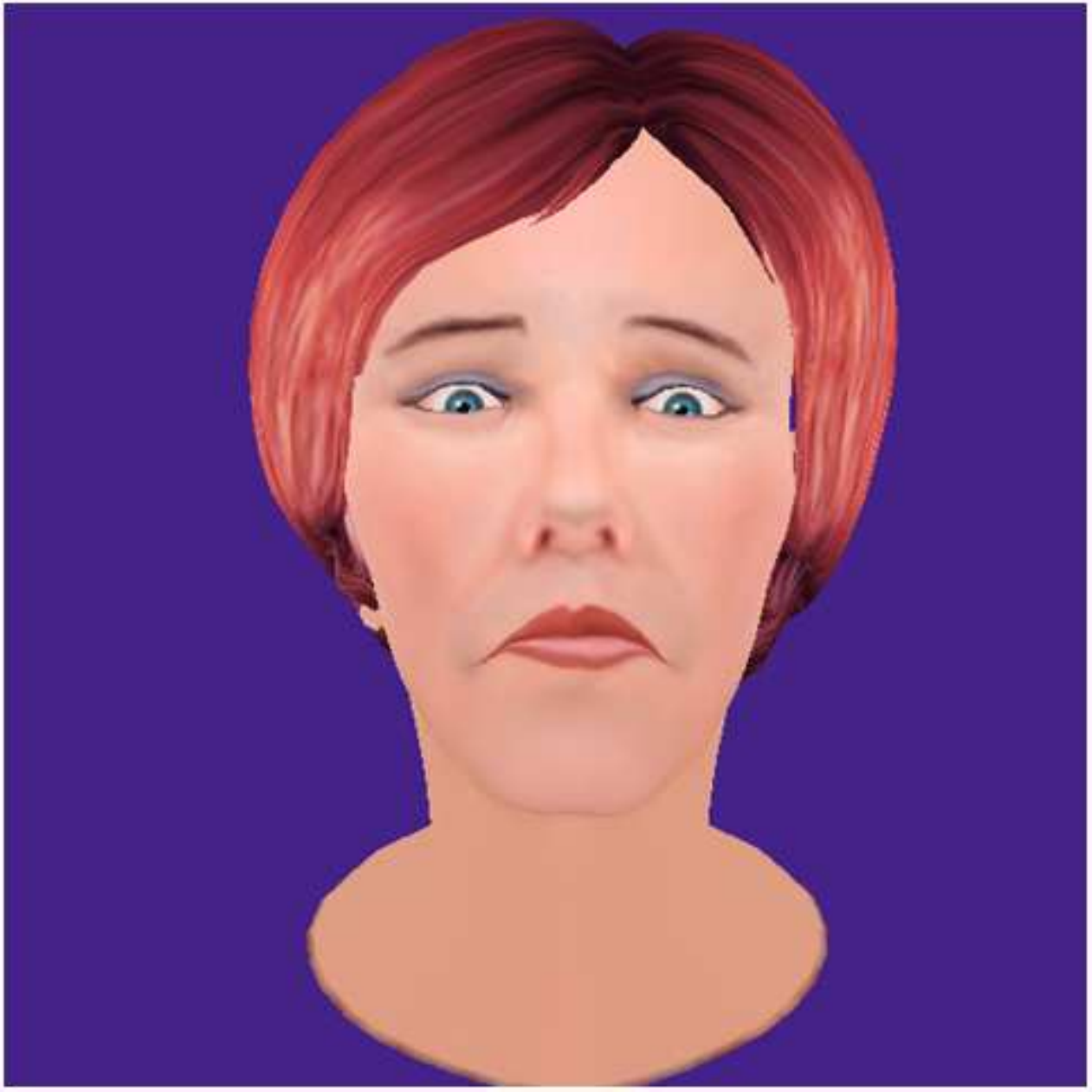}
}
\subfloat[][\label{surprise} Surprise]
{
	\centering
    \includegraphics[scale = 0.15]{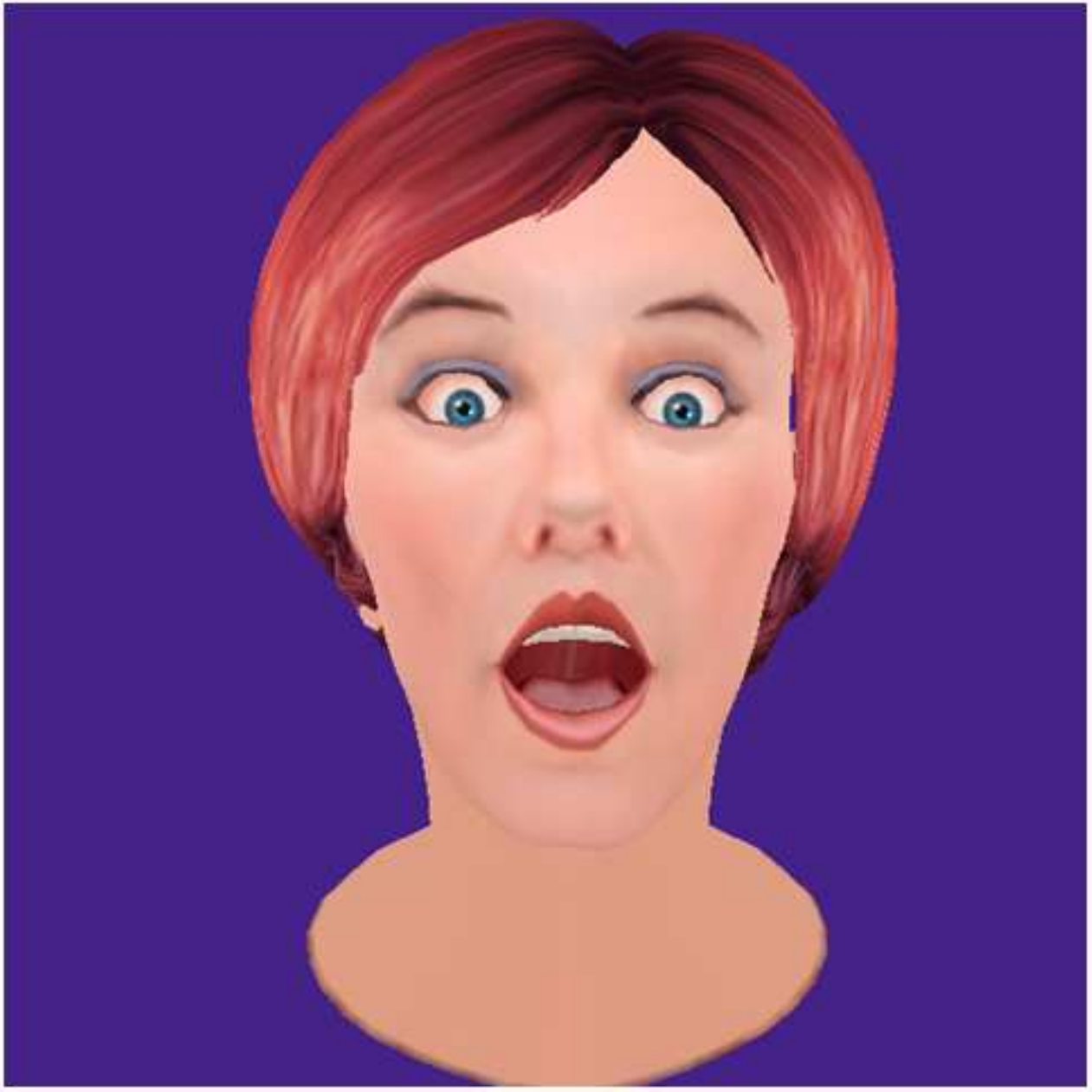}
}
\caption{\label{fig:SomeVisemesExpression}
Examples of some visemes and expressions
}
\vspace{-0.2cm}
\end{figure}

For a given language, visually similar phonemes are grouped into units called visemes. For example the consonants /b/, /p/ and /m/ in the words ``buy,'' ``pie,'' and ``my'' form a single viseme class. We categorized English phonemes into 20 viseme classes. These classes represent the articulation targets that lips and tongue move to during speech production. A graphic artist designed 3D models of these viseme classes in Maya. Figure~\ref{fig:SomeVisemesExpression}, demonstrates some visemes used in our animation system. Finally, natural visual speech is obtained by blending the proper models corresponding to each part of speech with different weights. 

To achieve a smooth and realistic look, we used a kernel smoothing technique. During speech production, the avatar system receives the time-aligned phonetic input from Bavieca system, converts the phonetic symbols into the corresponding visemes, which specifies the movements of the mouth and tongue, synchronized with the recorded or synthesized speech. The algorithm models coarticulation by smoothing across adjacent phonemes. 

Using the kernel technique resulted in smoother and more natural looking animations; however, when utterances included the labial phonemes /b/, /m/, /p/, which are accompanied by lip closure, the smoothing algorithm prevented the lips from closing when the duration of the labial phoneme is very short (e.g., 5 msec.) and the adjacent phoneme targets caused the lips to be open (e.g., /\textturnscripta/ as in ``mama''). To force lip closure for the labials, we extended the duration of labial visemes to include the closure interval (the period of relative silence before the sound is released, thus increasing the chance that at least one frame consisting of just the labial viseme will appear.

We designed the models in three portions: eyes, face and hair. This design allows them to be interchangeable and customizable, and gives us the ability to design any number of characters to easily change the robot's appearance. The system has the ability to control eye gaze independently of the visual speech and facial expression animation, and thus enables functionality to control eye gaze (e.g., in concert with face tracking).

\subsection{Lip Blending with Emotion}
\label{sec:Lip Blending with Emotion}
In order to blend the expressions with the lip movement, the animation uses the following formula to generate facial expressions based on the current viseme and emotion morph targets: 
\begin{equation}
\label{mouth}
F_{j}=F_{c}+ \lambda_j (F_j^{max} - F_0)
\end{equation}
where $F_{c}$ represents the current viseme, $F_j^{max}$ is the desired expression model at the maximum intensity, $F_0$ is the Neutral model. The parameter $\lambda_j \in [0,1]$ is the intensity of the $j^{th}$ expression model $F_{j}$. The graphic artist designed 3D models of six basic expressions (i.e., anger, disgust, fear, joy, sadness and surprise) in Maya based on Facial Action Coding System (FACS)~\cite{ekman1977facial}. For example joy involves Cheek Raiser (AU 6) and Lip Corner Puller (AU 12) and sadness involves Inner Brow Raiser (AU 1), Brow Lowerer (AU 4) and Lip Corner Depressor (AU 15).

In order to blend the expressions with the lip movement, adding weight to the emotion morph targets without regard to the movements of the face caused by speech production will result in unnatural looking facial expressions. For example, combining the surprise expression which causes the mouth to be fully open, conflicts with the production of phonemes like /b/, /f/ and /v/ that are produced with the lips closed or nearly closed. Combining the joy emotion with puckered mouth visemes such as /o/ will also result in visual speech and expressions that are not natural and are perceived as abnormal or creepy. To overcome this problem, we designed a table that provides a viseme weight factor and a maximum emotion weight for every viseme emotion combination. These values are adjusted empirically for each combination. 

We separated the facial expression morph targets into upper and lower face morph targets; the upper face includes everything from the tip of the nose upwards. The lower face includes the region below the nose; mainly the lips, lower cheeks and chin. This partitioning of the face enables us to adjust the weight of just the lower face morph target weights so that the upper face remains consistent with the morph targets of desired expressions. In addition, for labial and labiodental visemes (those for the letters m, b, p, f and v) that require the avatar's lips to be closed or nearly closed to look natural, we developed visemes pre-blended with the open mouthed emotions. These are used to replace the viseme and lower face expression when they come up in combination.

\subsection{Calibration}
\label{sec:Calibration}
\begin{figure}[h]
\centering
\subfloat[]
{
	\centering
    \includegraphics[scale = 0.17]{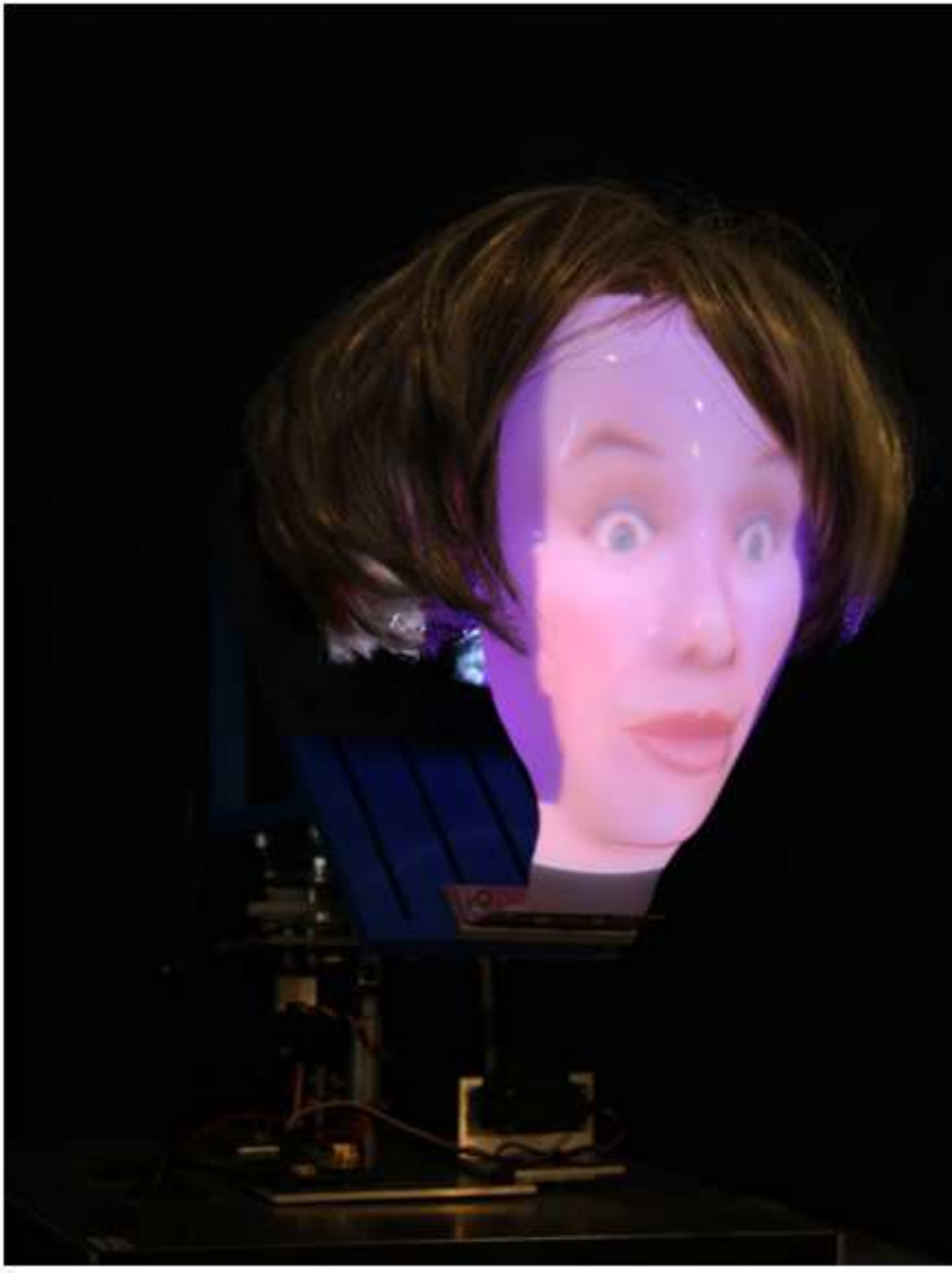}
}
\subfloat[]
{
    \includegraphics[scale = 0.17]{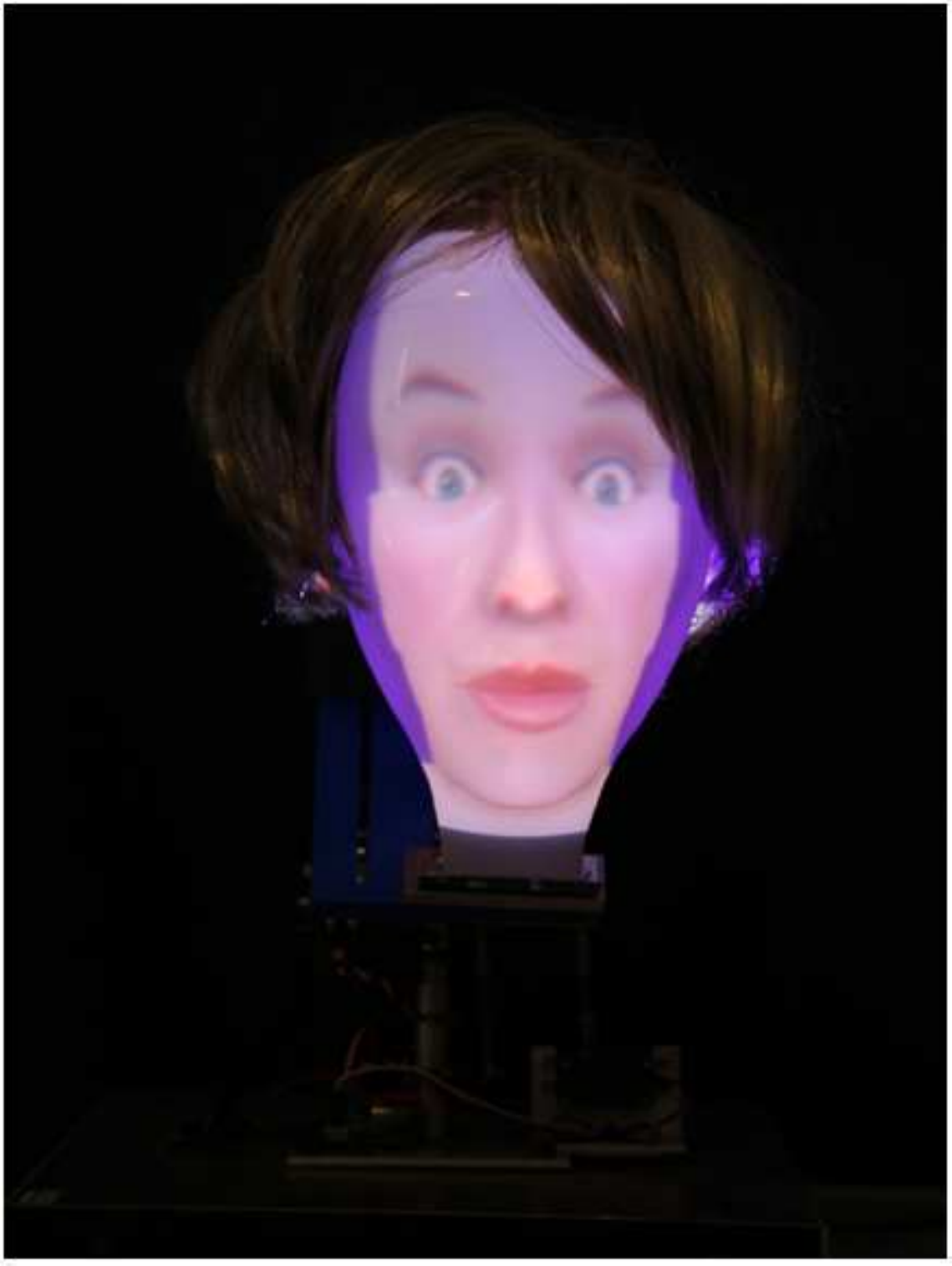}
}
\subfloat[]
{
	\centering
    \includegraphics[scale = 0.17]{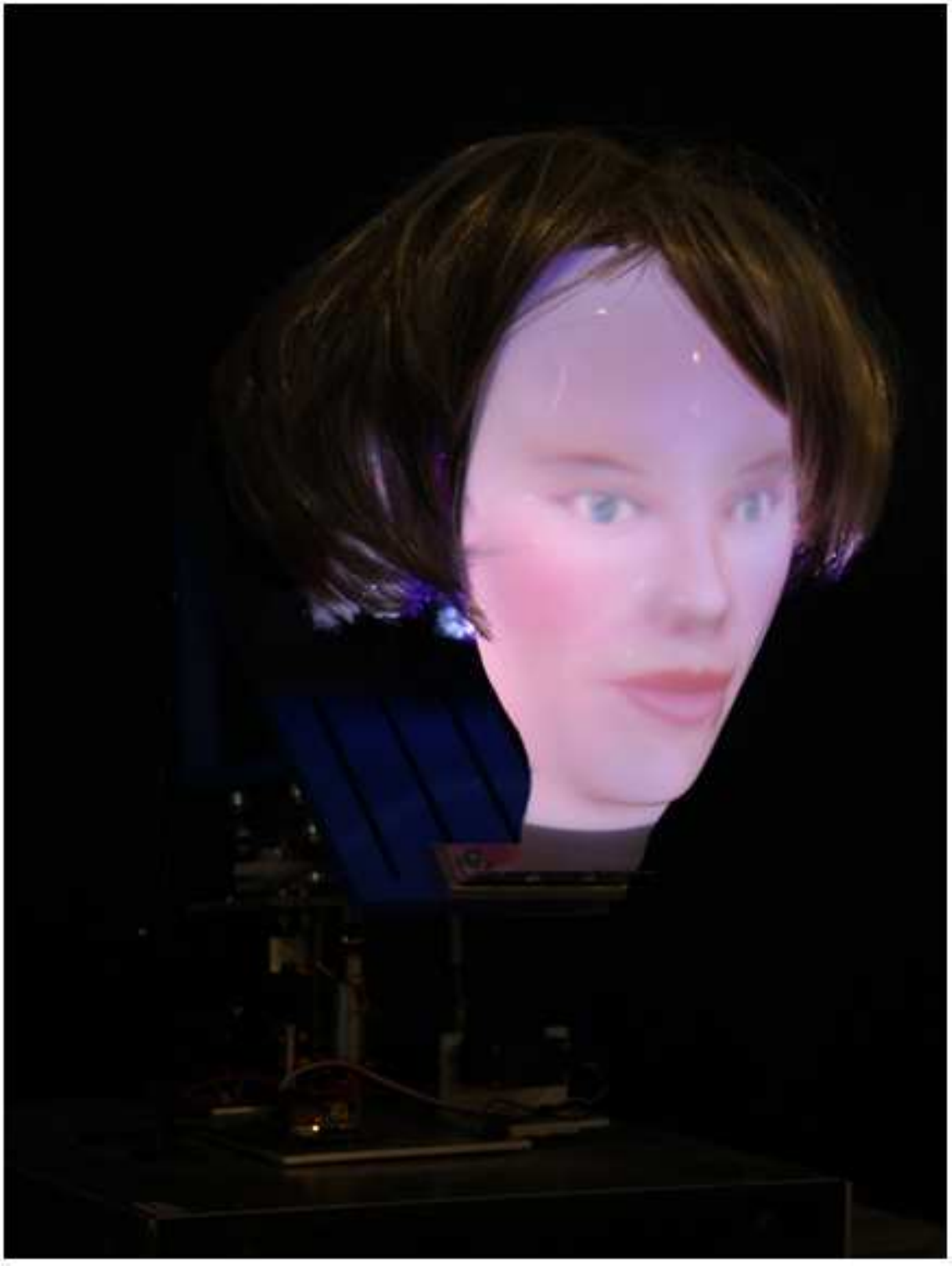}
}
\subfloat[]
{
    \includegraphics[scale = 0.17]{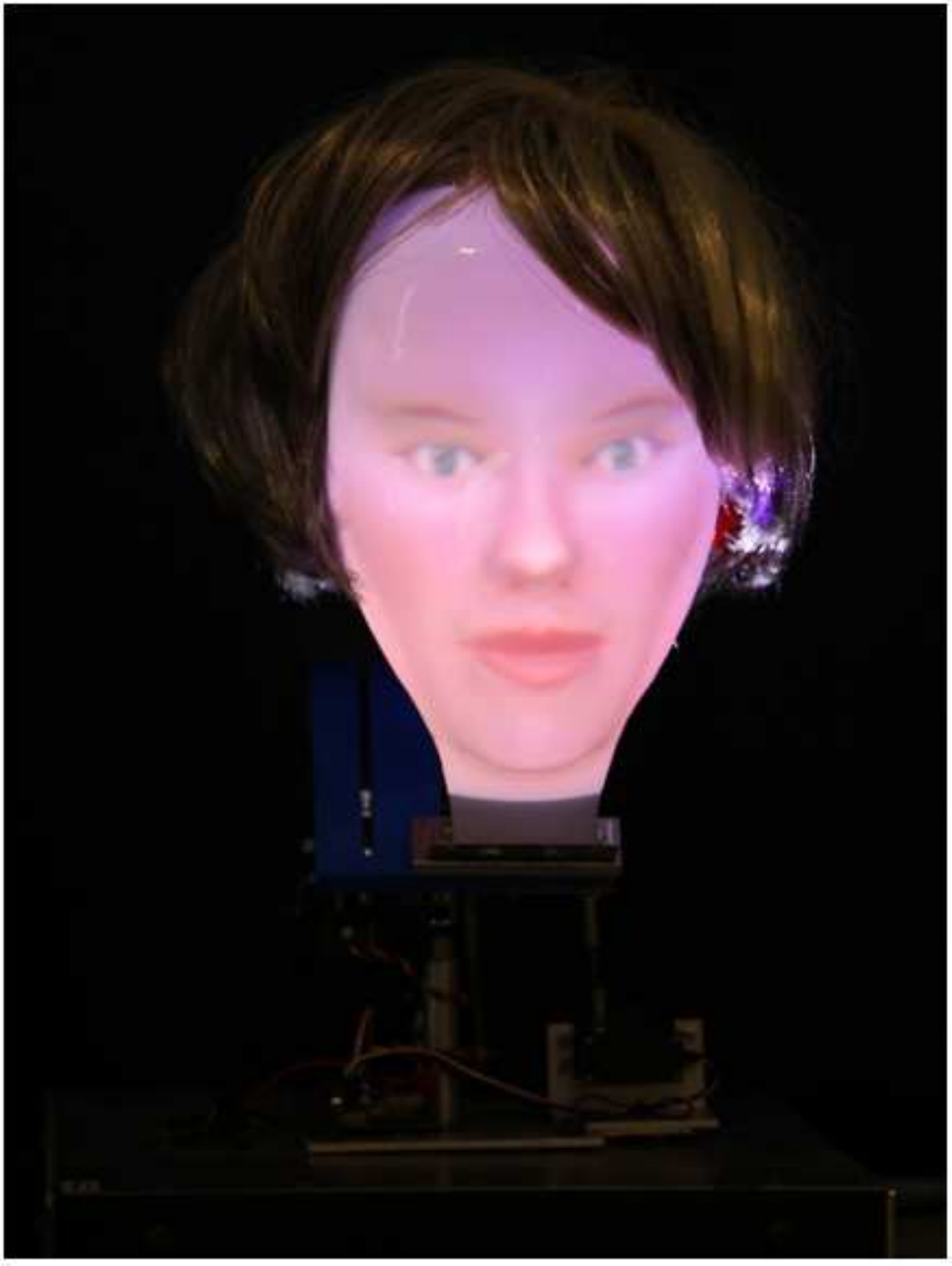}
}
\caption{\label{fig:ResultCalibration}
Result of calibration: (a) and (b) show projection without calibration, (c) and (d) show projection with calibration from side and frontal view.
}
\end{figure}

Due to the projector and fish eye lens distortions, the resulting direct projection of the animated face model on the mask appears distorted (See Fig.~\ref{fig:ResultCalibration}). Hence, we need to rectify the projection so the image appears undistorted and the facial regions (e.g., eyes, mouth) of the facial model are projected to the desired position on the mask. In order to achieve a smooth animation displaying at 30 fps, we decide to distort the original Maya models rather than rectifying the projection at run time in each frame. 

\begin{figure}[thpb]
  \centering
  \includegraphics[scale=0.235]{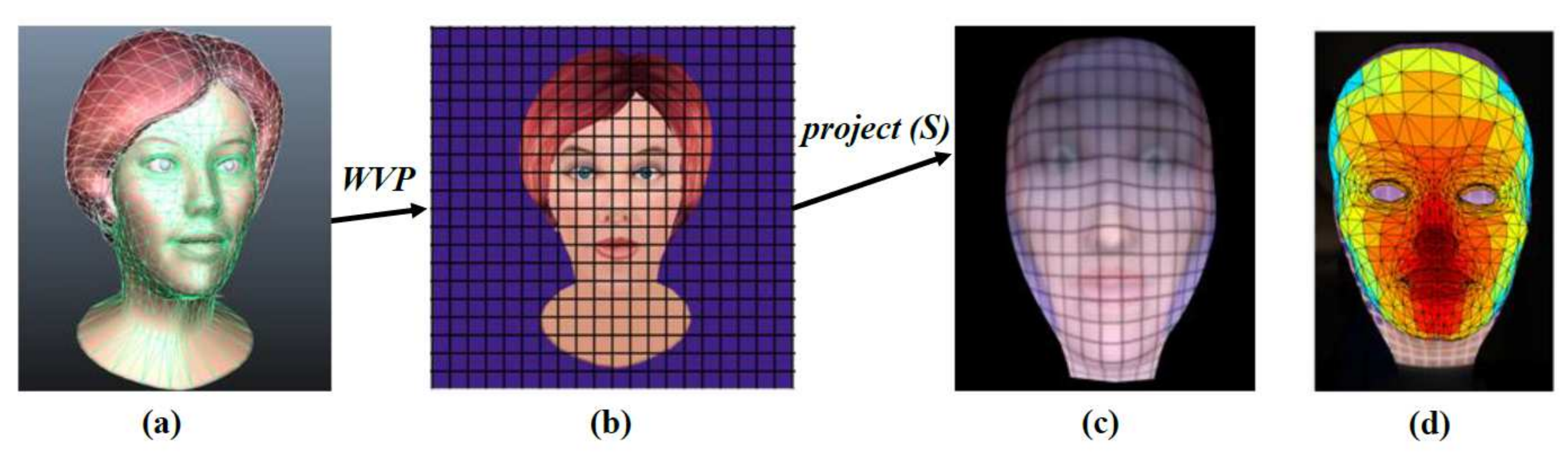}\\ 
  \vspace{-0.2cm}
  \caption{Calibration process.}
  \vspace{-0.3cm}
  \label{fig:CalibrationProcess}
\end{figure}

Assuming N is the neutral model in model coordinates (Fig.~\ref{fig:CalibrationProcess}.a), $S=N \times WVP$ is the displayed projection in the screen coordinates where $WVP$ matrix is the multiplication of the world, view and projection matrices, respectively (Fig.~\ref{fig:CalibrationProcess}.b). Assuming $M=project(S)$ is the projected model on the mask (Fig.~\ref{fig:CalibrationProcess}.c), we aim to find $N'$ (the distorted neutral model) such that $M'=project(S')$ looks undistorted on the mask, where $S'=N'×WVP$. In order to estimate the function $project(.)$, we create a checkerboard in the screen coordinates (Fig.~\ref{fig:CalibrationProcess}.b) and projected on the mask (Fig.~\ref{fig:CalibrationProcess}.c). Then, we define a piecewise homography mapping between the corresponding rectangles of the mask and the triangles displayed on the screen. To find the undistorted neutral model on the mask, $M'$, we apply an affine transformation to place the mold model, used to create the mask in the vacuum machine, on an image of the mask (Fig.~\ref{fig:CalibrationProcess}.d). Afterwards, we apply the piecewise homography on the mold model and replace the corresponding vertices of the neutral model in the screen coordinates, $S$, with distorted mold model to estimate $S'$. We finally use $N'=S'×WVP^{-1}$ as the neutral model in our application. Figure~\ref{fig:ResultCalibration} shows the results of our calibration. 

\section{HRI Experiments and Results}
\label{sec:HRI Experiments and Results}
In order to evaluate individuals' experiences and impressions of the ExpressionBot, we designed and conducted three experiments. Participants were 23 typical adults, 9 female and 14 males, with age range 18-51 years (Mean= 27.26, SD=7.79) and a variety of ethnicities (19 Caucasian, 2 Asian, 2 Hispanic). 

Hereafter, we refer to the 3D computer character on the computer screen as the screen-based agent, and the projection of the 3D model onto the robotic head as the physical agent. We used a 23'' LCD display to display the screen-based agent at the same size as the physical agent.

The objective of the first experiment was to assess how accurately subjects were able to interpret the projected facial expressions. Participants watched the robotic agent and the screen-based agent in two different sessions randomly (i.e. some participants observed the physical agent first while the others watched the screen-based agent first). A series of six basic emotions (joy, sadness, surprise, disgust, fear and anger) were displayed in random order. Each expression was displayed one time for about 5 seconds.  The subject was then asked to select one of the six categories. They could also respond ``none,'' if they were unable to assign the facial expression to one of the six categories. 

Tables~\ref{tbl:Confusion_physical_agent} and~\ref{tbl:Confusion_screen-based_agent} present confusion matrices of the intended and classified expressions displayed on the physical agent and the screen-based agent, respectively. Comparing the percentages reported in these tables shows that the surprise and sad emotions were recognized perfectly (100\% recognition rate) in both agents by the participants. The joy emotion was recognized perfectly when displayed on the physical agent but was recognized 92\% of the time on the screen-based agent.  Interestingly, Anger was recognized correctly 85\% of the time for the physical agent, and only 38\% of the time for the screen-based agent. Disgust was classified as anger more often than it was classified as disgust for both agents, and fear was recognized correctly over 50\% of the time in both agents, and confused most often with sadness.  In sum, the results showed high recognition rates for Joy, Sadness and Surprise in both agents, lower and similar recognition rates for Disgust and Fear in the two agents, and superior performance for Anger when displayed on the physical agent.

\begin{table}[htpb]
\caption{Confusion matrix of recognized expression on the physical agent}
\label{tbl:Confusion_physical_agent}
\vspace{-0.3cm}
\centering
\begin{tabular}{|c|c|c|c|c|c|c|c|}
\hline
\scriptsize{\%} & \scriptsize{Joy}&\scriptsize{Anger}&\scriptsize{Sadness}&\scriptsize{Disgust}&\scriptsize{Surprise}&\scriptsize{Fear}&\scriptsize{None} \\
\hline
\scriptsize{Joy} & \scriptsize{\textbf{100}} & \scriptsize{0} & \scriptsize{0} & \scriptsize{0} & \scriptsize{0} & \scriptsize{0} & \scriptsize{0} \\
\hline
\scriptsize{Anger} & \scriptsize{0} & \scriptsize{\textbf{85}} & \scriptsize{0} & \scriptsize{10} & \scriptsize{0} & \scriptsize{0} & \scriptsize{5} \\
\hline
\scriptsize{Sadness} & \scriptsize{0} & \scriptsize{0} & \scriptsize{\textbf{100}} & \scriptsize{0} & \scriptsize{0} & \scriptsize{0} & \scriptsize{0} \\
\hline
\scriptsize{Disgust} & \scriptsize{0} & \scriptsize{60} & \scriptsize{0} & \scriptsize{\textbf{40}} & \scriptsize{0} & \scriptsize{0} & \scriptsize{0} \\
\hline
\scriptsize{Surprise} & \scriptsize{0} & \scriptsize{0} & \scriptsize{0} & \scriptsize{0} & \scriptsize{\textbf{100}} & \scriptsize{0} & \scriptsize{0} \\
\hline
\scriptsize{Fear} & \scriptsize{0} & \scriptsize{0} & \scriptsize{35} & \scriptsize{10} & \scriptsize{0} & \scriptsize{\textbf{55}} & \scriptsize{0} \\
\hline
\end{tabular}
\end{table}

\vspace{-0.4cm}
\begin{table}[htpb]
\caption{Confusion matrix of recognized expression on the screen-based agent}
\label{tbl:Confusion_screen-based_agent}
\vspace{-0.3cm}
\centering
\begin{tabular}{|c|c|c|c|c|c|c|c|}
\hline
\scriptsize{\%} & \scriptsize{Joy}&\scriptsize{Anger}&\scriptsize{Sadness}&\scriptsize{Disgust}&\scriptsize{Surprise}&\scriptsize{Fear}&\scriptsize{None} \\
\hline
\scriptsize{Joy} & \scriptsize{\textbf{92}} & \scriptsize{0} & \scriptsize{0} & \scriptsize{8} & \scriptsize{0} & \scriptsize{0} & \scriptsize{0} \\
\hline
\scriptsize{Anger} & \scriptsize{0} & \scriptsize{\textbf{38}} & \scriptsize{8} & \scriptsize{46} & \scriptsize{0} & \scriptsize{8} & \scriptsize{5} \\
\hline
\scriptsize{Sadness} & \scriptsize{0} & \scriptsize{0} & \scriptsize{\textbf{100}} & \scriptsize{0} & \scriptsize{0} & \scriptsize{0} & \scriptsize{0} \\
\hline
\scriptsize{Disgust} & \scriptsize{8} & \scriptsize{46} & \scriptsize{0} & \scriptsize{\textbf{38}} & \scriptsize{8} & \scriptsize{0} & \scriptsize{0} \\
\hline
\scriptsize{Surprise} & \scriptsize{0} & \scriptsize{0} & \scriptsize{0} & \scriptsize{0} & \scriptsize{\textbf{100}} & \scriptsize{0} & \scriptsize{0} \\
\hline
\scriptsize{Fear} & \scriptsize{0} & \scriptsize{0} & \scriptsize{38} & \scriptsize{0} & \scriptsize{8} & \scriptsize{\textbf{54}} & \scriptsize{0} \\
\hline
\end{tabular}
\end{table}

In the second experiment, we evaluated the proposed method for visual speech and examined subjects' judgments of speech production quality using the physical agent. Two short segments of speech were used in this experiment. Segment 1 was a seven second interval of a Margaret Thatcher's speech with length of 11 seconds while segment 2 was a seven second interval of Microsoft Anna synthetic speech. We chose these segments to cover a variety of length, speed, accent, and different phonemes (vowels, consonants and labial phonemes). Each speech was played two times with different lip synchronization approaches: 1) A basic approach where at each phoneme only the corresponding viseme was displayed without any kernel smoothing; 2) The proposed approach described in Section~\ref{sec:Animation} where lip closure was enforced in labial phonemes and kernel smoothing was applied.

We asked the participants to rate how realistic the visual speech looked on a scale from 0 to 5, where 0 being unrealistic and 5 being very realistic. Table~\ref{tbl:VisualSpeechRating} shows the evaluation of the two speech segments for the two different visual speech approaches displayed on the physical agent and screen-based agent. One-tail paired T-test analyses were conducted, where the results show that there was not significant preferences between the physical and on-screen agents. However, the T-Test analysis indicated a significant preference for the proposed approach for synchronizing lips with speech over basic approach (p=.001 and p=.0002 on the physical agent and p=.0933 and p=.0067 on the screen-based agent for the speech segments 1 and 2, respectively).

\begin{table}[h]
\caption{Average (STD) values of visual speech rating on the physical agent and the screen-based agent}
\label{tbl:VisualSpeechRating}
\vspace{-0.3cm}
\begin{center}
\begin{tabular}{l|c|c|c|c|}
\cline{2-5}
\multirow{2}{*}{}                       & \multicolumn{2}{c|}{\textbf{Physical Agent}} & \multicolumn{2}{c|}{\textbf{Screen-based Agent}} \\ \cline{2-5} 
                                        & Basic                 & Proposed             & Basic                   & Proposed               \\ \hline
\multicolumn{1}{|l|}{\textbf{Speech 1}} & 3.04 (0.80)           & 3.85 (0.85)          & 3.06 (0.79)             & 3.53 (0.91)            \\ \hline
\multicolumn{1}{|l|}{\textbf{Speech 2}} & 2.50 (0.88)           & 3.45 (0.75)          & 2.42 (0.93)             & 3.21 (0.69)            \\ \hline
\end{tabular}
\end{center} 
\vspace{-0.15cm}
\end{table}

In~\cite{ma_animating_2004}, the Mona Lisa effect on an animation displayed on a 2D screen and its 3D projection on a head model was studied by shifting eye gaze every 2 degrees from left to right. Then the participants were asked to report their perception of the agents' eye gaze direction. Their results showed a clear Mona Lisa effect in the 2D setting and many subjects perceived a mutual gaze with the screen-based agent's head at the same time for frontal and near frontal gaze angles.  For the physical agent, the Mona Lisa effect was completely eliminated.

\begin{figure}[thpb]
  \centering
  \includegraphics[scale=0.22]{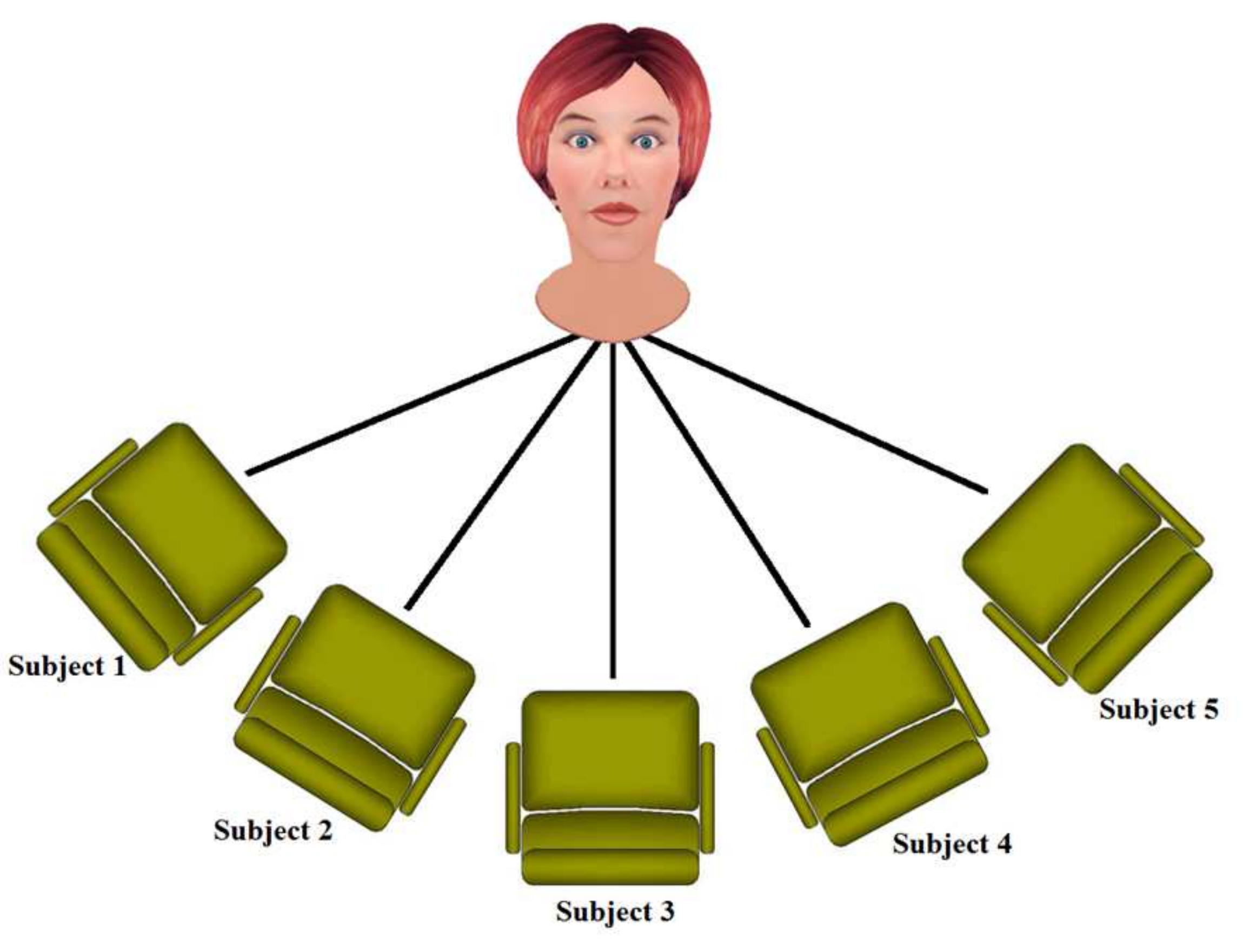}\\ 
  \caption{Experimental setup and placement of subjects.}
  \label{fig:ExperimentalSetup}
\end{figure}

Inspired by the experimental setting in~\cite{ma_animating_2004}, we evaluated the perception of the eye gaze direction of the physical agent and screen-based agent. In this experiment, five subjects were simultaneously seated around an animated agent in two separate sessions; one session to examine the screen-based agent and another session for the physical agent (see Fig.~\ref{fig:ExperimentalSetup}). The seats were positioned at -45$^{\circ}$, -25$^{\circ}$, 0$^{\circ}$, 25$^{\circ}$, 45$^{\circ}$, where 0$^{\circ}$ is the seat in front of the agent. The distance of the subjects to the agent was five feet. 

In the first setting (called eye gaze only), the agent's head looked straight and only the eye gaze was shifted towards each subject. After each shift, all the subjects were asked to report the subject number that the agent was looking at. We repeated each experiment three times with 10 random eye gazes each time. In this setting, the subjects perceived the direction of the eye gaze 50\% (SD=24\%) of the times correctly for the screen-based agent and 88\% (SD=13\%) of the times correctly for the physical agent (p$<$0.005).

In the second setting (called eye gaze plus arbitrary head movement), the agent rotated its head and shifted its eye gaze randomly at the same time, but the head was not necessarily towards the subject of interest. Then, after each head movement and gaze shift, all the subjects reported the subject number that the agent was looking at. We repeated each experiment three times with 10 random eye gazes each time. In this setting, subjects perceived the direction of the eye gaze 43\% (SD=18\%) of the time correctly for the screen-based agent and 77\% (SD=15\%) of the times correctly for the physical agent (p$<$.000017). These results show that compared with the screen-based agent, the subjects perceived the eye gaze direction produced by the robotic face more accurately in both the ``eye gaze only'' and ``eye gaze plus arbitrary head movement'' settings.

\section{Conclusion}
\label{sec:Conclusion}
This article described the design and creation of a low-cost emotive robotic head, called ExpressionBot, for natural face-to-face communication (the cost of the hardware system is about \$1500). The results of our HRI studies on a group of participants illustrated that the subjects perceived the facial expression anger with a much greater accuracy in the robotic face than the screen-based face and they also rated the generated visual speech smooth and realistic on both robotic and screen-based systems. In addition, we studied the perception of eye gaze's direction in two experiments, one in which the head was frontal and only the eye gaze was shifted, and the other with the head rotated but not necessarily correlated with the eye gaze direction. In both experiments, our results showed that participants perceived the robotic face mutual gaze more accurately. 

The developed robotic head represents a new level of integration of emotive capabilities that enables researchers to study socially emotive robots/agents that can generate spoken-language, show emotions, and communicate effectively with people in a natural way as humans do. Such systems can be applied in many domains including health-care, education, entertainment, and home-care. It will also be an ideal platform for designing a new generation of more immersive and effective intelligent tutoring and therapy systems, and robot-assisted therapeutic treatments.

\bibliographystyle{IEEEtran}

\end{document}